\documentclass{article}

\usepackage[preprint]{corl_2026} 

\title{RT-VLA: Real-Time Vision-Language-Action Models via Knowledge Distillation}

\author{
  Xiangyu Huang\thanks{Equal contribution.} \\
  Carnegie Mellon University \\
  xiangyuh@andrew.cmu.edu \\
  \And
  Zhenlin Hua\footnotemark[1]\\
  Carnegie Mellon University \\
  zhenlinh@andrew.cmu.edu \\
  \And
  Han Zhou \\
  Carnegie Mellon University \\
  hanzhou2@andrew.cmu.edu \\
  \And
  Shounak Sural \\
  Carnegie Mellon University \\
  ssural@andrew.cmu.edu \\
  \And
  Ragunathan (Raj) Rajkumar \\
  Carnegie Mellon University \\
  rajkumar@andrew.cmu.edu 
}

\usepackage{amsmath}
\usepackage{amssymb}
\usepackage{booktabs}
\usepackage{graphicx}
\usepackage{subcaption}
\usepackage{multirow}
\usepackage{titlesec}

\titlespacing*{\section}
{0pt}{0.45em}{0.25em}

\titlespacing*{\subsection}
{0pt}{0.3em}{0.15em}

\begin{document}
\maketitle
\raggedbottom


\begin{abstract}
Vision-Language-Action (VLA) models have shown strong potential for end-to-end autonomous driving by jointly modeling visual perception, language reasoning, explainability and action prediction. However, their large vision-language backbones and reasoning modules introduce substantial inference latency and thereby prevent their deployment in the unforgiving reality of the road networks. We propose RT-VLA, a lightweight, distilled VLA model that transfers the driving and reasoning capabilities of the state-of-the-art SimLingo model into a compact student through multi-level supervised distillation. RT-VLA preserves language-based reasoning and supports post-hoc explanation through offline language analysis of safety-critical driving moments without adding latency to real-time control. Compared to the SimLingo teacher, RT-VLA maintains competitive closed-loop driving and language reasoning performance while reducing inference time by \textbf{$44.8\times$} in vision-only mode and \textbf{$7.9\times$} in vision+language mode. These results suggest that supervised distillation is a practical approach for building real-time, explainable VLA-style autonomous driving models. 
\end{abstract}


\vspace{-2mm}
\section{Introduction}
\vspace{-1mm}
End-to-end (E2E) autonomous driving maps sensory observations directly to driving actions, avoiding hand-designed decomposition into perception, prediction, planning and control. 
Recent Vision-Language-Action (VLA) models extend this paradigm by integrating visual perception, language reasoning and action prediction within a unified framework. 
Representative methods, including DriveCoT~\cite{drivecot2024}, AutoVLA~\cite{autovla2025}, OpenDriveVLA~\cite{opendrivevla2026}, CarLLaVA~\cite{carllava2024}, ORION~\cite{orion2025}, Alpamayo~\cite{wang2025alpamayo} and SimLingo~\cite{simlingo2025}, demonstrate that language can support high-level reasoning, instruction following and interpretable decision making. 
However, these benefits come with substantial inference latency due to large vision-language backbones, autoregressive generation and explicit reasoning modules.

The latencies and computational demands of VLA models limit deployment in any public transportation corridors. 
In dense traffic and constrained urban environments, the ego vehicle must quickly respond to surrounding vehicles, pedestrians, traffic signals and sudden road changes. 
Slow inferencing leads to unsafe delays in trajectory updates, reduced responsiveness and inefficient decision making in time-sensitive contexts. 
Therefore, a practical VLA-style driving model should ideally preserve safe driving and reasoning capabilities while reducing inference cost.

In this paper, we propose \textbf{RT-VLA}, a real-time distilled VLA model for E2E autonomous driving. 
RT-VLA transfers the driving and reasoning capabilities of a larger SimLingo teacher into a compact student model through multi-level supervised distillation over visual features, query representations, waypoint predictions and language logits. 
RT-VLA supports offline post-hoc explanation by logging driving observations from the efficient driving branch and invoking a posteriori language reasoning branch only when explainability is needed. 

Our contributions are summarized as follows:
\begin{itemize}
    \item We propose RT-VLA, a lightweight distilled VLA model that preserves driving and language-based reasoning capabilities while substantially reducing inference latency.
    \item Unlike existing distillation methods, we design a multi-level distillation strategy that transfers teacher knowledge through visual features, query representations, waypoint predictions and language logits.
    \item We introduce efficient reasoning via language-logit distillation and on-policy language fine-tuning, enabling offline post-hoc explanation without adding runtime latency.
    \item We show a favorable performance-efficiency trade-off on Bench2Drive \cite{jia2024bench2drive}, achieving driving and language scores comparable to SimLingo while reducing inference time by \textbf{$44.8\times$} in vision-only mode and \textbf{$7.9\times$} in vision+language mode.
    
\end{itemize}

\vspace{-2mm}
\section{Related Work}
\label{sec:related_work}
\vspace{-1mm}
    \subsection{End-to-End Autonomous Driving}
    \vspace{-1mm}
    End-to-end (E2E) autonomous driving learns a direct mapping from sensory observations to driving decisions, such as trajectories, waypoints or control commands, without relying on a fully hand-designed modular stack. 
    Compared with traditional pipelines that separately handle perception, prediction, planning and control, E2E approaches may reduce system design time and allow the driving policy to be optimized jointly for closed-loop behavior. 
    Early imitation-learning methods such as Conditional Imitation Learning (CIL)~\cite{cil2018} showed that visual policies can follow different driving intents when conditioned on high-level navigational commands. 
    Learning by Cheating (LBC)~\cite{lbc2020} further demonstrated privileged supervision by training an agent with privileged state information and distilling its behavior into a vision-based policy. 
    Later methods improved E2E driving with more capable architectures and richer representations: TCP~\cite{tcp2022} jointly predicts trajectories and controls in a camera-only framework, while TransFuser~\cite{transfuser2023} and InterFuser~\cite{interfuser2023} introduce transformer-based sensor fusion and interpretable representations for urban driving. 
    Despite their effectiveness, conventional E2E policies remain limited in interpretability, causal reasoning and long-tail robustness because decisions are typically represented only as trajectories, waypoints or low-level controls.
    \vspace{-1mm}
    \subsection{VLA Models for Autonomous Driving}
    \vspace{-1mm}
    With advances in large language models and multimodal learning, Vision-Language-Action (VLA) models have become a promising paradigm for end-to-end autonomous driving. 
    These models extend vision-language models with action generation, allowing visual observations, language instructions and driving contexts to be mapped into executable behaviors. 
    Compared with conventional vision-action policies, VLA models use language as an intermediate reasoning interface, enabling semantic scene understanding, instruction following and interpretable explanations. 
    DriveCoT~\cite{drivecot2024} introduces chain-of-thought supervision for end-to-end driving, showing that structured reasoning labels can improve interpretability and decision-making. 
    Recent VLA frameworks, including ORION~\cite{orion2025}, AutoVLA~\cite{autovla2025} and OpenDriveVLA~\cite{opendrivevla2026}, integrate language-based reasoning with trajectory or action generation to improve planning and instruction following. 
    
    CarLLaVA~\cite{carllava2024}, the basis of SimLingo-BASE, integrates vision-language modeling into camera-only closed-loop driving and predicts geometric route and temporal speed waypoints. 
    SimLingo~\cite{simlingo2025} extends it into a full VLA model by adding vision-language understanding, commentary generation and language-action alignment through \emph{Action Dreaming}. 
    While both achieve strong driving and reasoning performance, their large vision-language backbones and autoregressive reasoning components introduce substantial latency. 
    Our work addresses this core limitation by distilling SimLingo's driving and reasoning capabilities into a lightweight student for real-time real-world deployment.
\vspace{-1mm}
    \subsection{Knowledge Distillation for Efficient Driving Models}
\vspace{-1mm}
    Knowledge distillation transfers knowledge from a large teacher model to a smaller student model, enabling efficient inference while preserving task performance~\cite{hinton2015distilling}. 
    Early distillation methods primarily supervise the student using the teacher's output distribution, while later approaches such as FitNets~\cite{fitnets2015} show that intermediate representations can also provide effective supervision for training compact models. 
    This idea is highly relevant for autonomous driving, where real-time closed-loop deployment requires accurate trajectory prediction and low inference latency. 
    In driving, privileged-teacher learning has been explored by Learning by Cheating~\cite{lbc2020}, which trains an agent with privileged state information and uses it to supervise a vision-based student policy. 
    On-policy imitation methods such as DAgger~\cite{dagger2011} address distribution mismatch by collecting supervision on states visited by the learned policy rather than only expert trajectories.
    Unlike prior distillation approaches that focus mainly on action prediction, our method specifically preserves language-related commentary capability while decoupling expensive language analysis from real-time planning.

\vspace{-2mm}
\section{Methodology}
\vspace{-2mm}
    We propose RT-VLA, a knowledge distillation framework~\cite{hinton2015distilling} for training a lightweight and efficient Vision-Language-Action (VLA) model for autonomous driving.
    \vspace{-1mm}
    \subsection{Overall Architecture}
\begin{figure*}[t]
      \centering
      \includegraphics[width=0.98\textwidth]{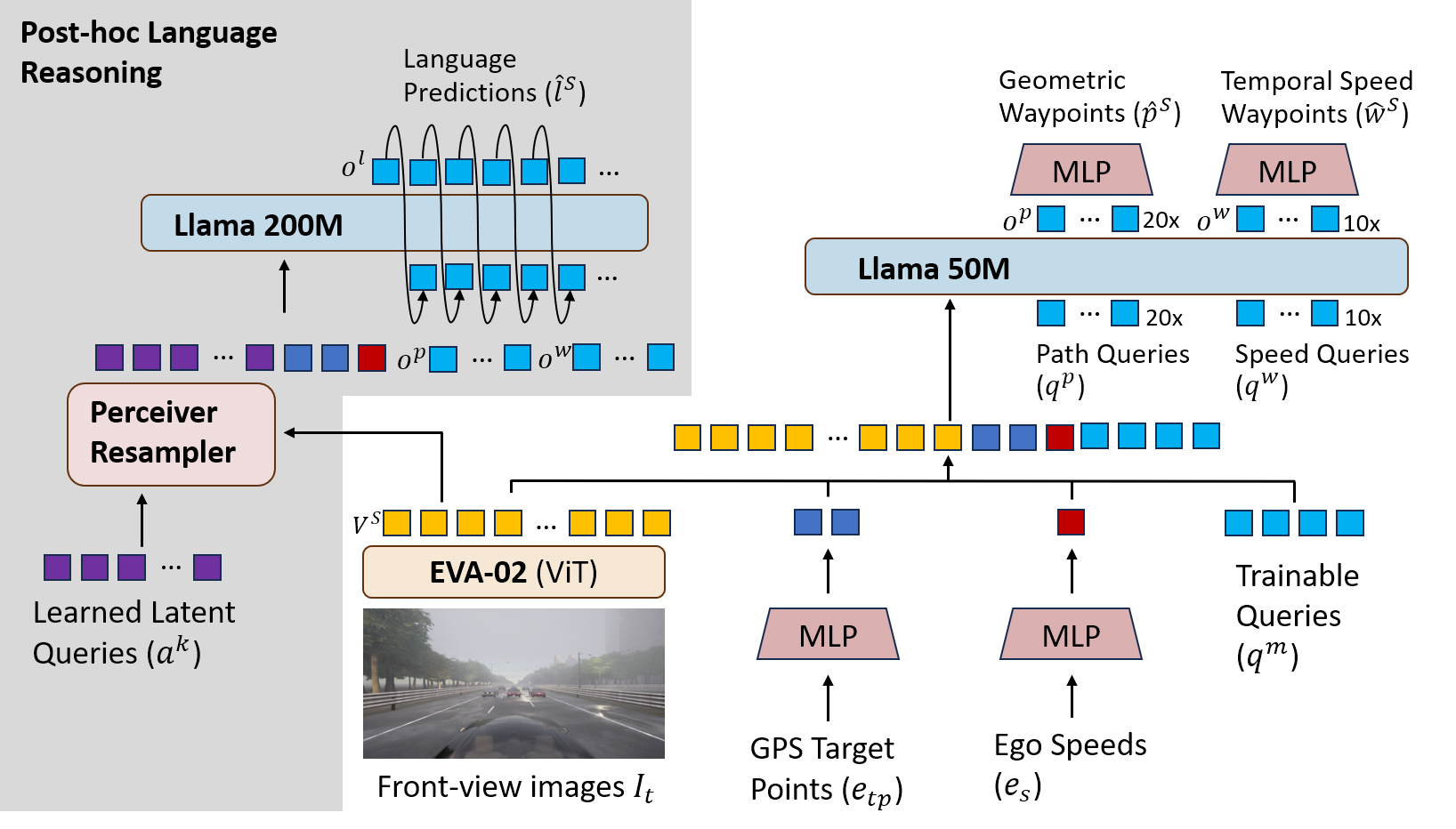}
      \caption{
      Overview of the RT-VLA student architecture.
      }
      \label{fig:student_architecture}
    \end{figure*}

    \vspace{-1mm}
    Our framework consists of a frozen teacher model and a trainable student model. 
    The teacher follows the SimLingo-style VLA architecture~\cite{simlingo2025}, using an InternVL-2 vision encoder~\cite{internvl} to extract high-capacity visual representations and a Qwen2-0.5B language model to produce driving and language outputs. 
    To construct an efficient student model, we use EVA-02~\cite{eva} as the visual encoder and lightweight language modules for driving prediction and language reasoning, as shown in Fig.~\ref{fig:student_architecture}.
    
    Given a front-view camera image $\mathbf{I}_t$, the student vision encoder produces visual tokens $\mathbf{V}^{S}$. 
    The ego speed and GPS target points are embedded by lightweight MLPs into state tokens, denoted as $\mathbf{e}_{s}$ and $\mathbf{e}_{tp}$, respectively. 
    These tokens are concatenated with trainable query tokens $\mathbf{q}^{m}$, path queries $\mathbf{q}^{p}$ and speed queries $\mathbf{q}^{w}$ and are processed by a lightweight driving language model. 
    The trainable query tokens $\mathbf{q}^{m}$ are learned embeddings that carry task-relevant information and provide additional capacity for aggregating driving context. 
    The output query representations are then passed to prediction heads to generate geometric waypoints $\hat{\mathbf{p}}^{S}$ and temporal speed waypoints $\hat{\mathbf{w}}^{S}$.
    
\begin{figure*}[t]
      \centering
      \includegraphics[width=0.98\textwidth]{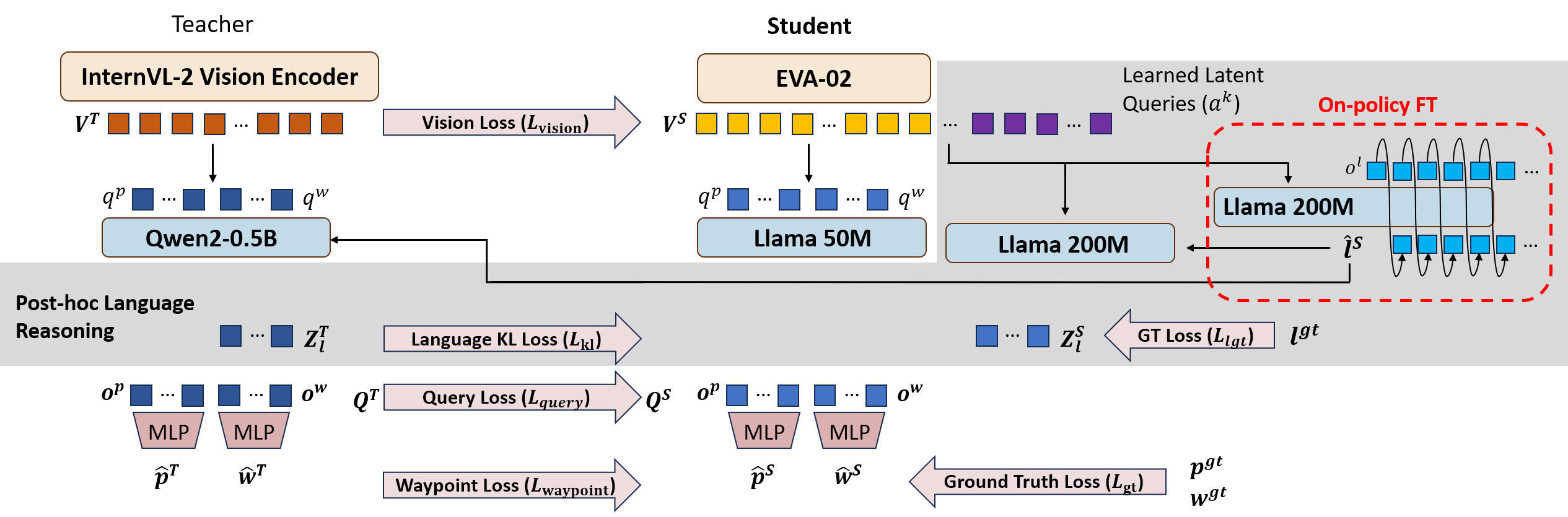}
      \caption{
      Multi-level distillation framework with on-policy fine-tuning.
      }
      \label{fig:distillation_architecture}
    \end{figure*}

    RT-VLA, in addition to its real-time driving branch, contains a lightweight language reasoning branch. 
    Visual tokens are compressed through a Perceiver Resampler ~\cite{alayrac2022flamingo} using learned latent queries $\mathbf{a}^{k}$ and processed by a compact language model to produce language logits $\mathbf{Z}_{\ell}^{S}$ and language predictions $\hat{\boldsymbol{\ell}}^{S}$.
    During language generation, key-value (KV) caching ~\cite{pope2023efficiently} reuses previously computed attention states across decoding steps, reducing the autoregressive generation overhead of the language reasoning branch.
    This branch supports reasoning and post-hoc explanation and is decoupled from the real-time driving branch during closed-loop control to avoid significant inference latency.
    During training, the teacher model is kept frozen. Fig.~\ref{fig:distillation_architecture} shows how the teacher model provides visual features, query representations, waypoint predictions and language logits to supervise the student model via multi-level distillation and on-policy fine-tuning (details in Sec. \ref{on-policy}).
    
    \vspace{-1mm}
    \subsection{Vision Encoder}
    \vspace{-1mm}
    The teacher and student models process visual observations with different model capacities. 
    The teacher extracts high-dimensional visual features $\mathbf{V}^{T} \in \mathbb{R}^{N_t \times D_t}$ using a large InternVL-2 vision encoder, while the student processes the front-view image with EVA-02 to obtain features $\mathbf{V}^{S} \in \mathbb{R}^{N_s \times D_s}$. 
    EVA-02 is adopted as the student vision encoder because its masked image modeling pre-training provides dense visual representations while remaining more efficient than the teacher's. 
    
    Since the teacher and student models have different feature dimensions and token lengths, direct feature matching is not possible. 
    We therefore introduce an alignment module that projects the student feature dimension to the teacher feature dimension and resizes the teacher token sequence to the student token length. 
    The aligned features are then used for visual feature distillation, encouraging the lightweight student to recover informative visual representations from a compact input stream.
    \vspace{-1mm}
    \subsection{Loss Training}
    \label{loss_des}
    \vspace{-1mm}
    The student model is trained with waypoint and language ground-truth supervision, together with distillation over visual features, query representations, waypoint predictions and language logits.
    
    Let $S$ and $T$ denote the student and teacher model respectively. 
    For each training sample, the student predicts geometric waypoints 
    $\hat{\mathbf{p}}^{S} \in \mathbb{R}^{H_p^S \times d_p}$ and temporal speed waypoints 
    $\hat{\mathbf{w}}^{S} \in \mathbb{R}^{H_w^S \times d_w}$, where $H_p^S$ and $H_w^S$ are the student prediction horizons for these waypoints. 
    The corresponding ground-truth targets are denoted as $\mathbf{p}^{gt}$ and $\mathbf{w}^{gt}$. 
    The teacher produces waypoint predictions 
    $\hat{\mathbf{p}}^{T} \in \mathbb{R}^{H_p^T \times d_p}$ and 
    $\hat{\mathbf{w}}^{T} \in \mathbb{R}^{H_w^T \times d_w}$.
    The teacher and student also produce visual features 
    $\mathbf{V}^{T} \in \mathbb{R}^{N_t \times D_t}$ and 
    $\mathbf{V}^{S} \in \mathbb{R}^{N_s \times D_s}$, query features
    $\mathbf{Q}^{T} \in \mathbb{R}^{M_t \times D_t}$ and 
    $\mathbf{Q}^{S} \in \mathbb{R}^{M_s \times D_s}$ and language logits
    $\mathbf{Z}_{\ell}^{T}$ and $\mathbf{Z}_{\ell}^{S}$.

    Because the teacher and student models have different feature dimensions and sequence lengths, alignment operators are used for feature and prediction matching. 
    Let $\phi_v(\cdot)$ and $\phi_q(\cdot)$ be learnable projections that map student features to the teacher feature dimension and let $\mathcal{A}_{N_s}(\cdot)$, $\mathcal{A}_{M_s}(\cdot)$, $\mathcal{A}_{H_p^S}(\cdot)$ and $\mathcal{A}_{H_w^S}(\cdot)$ denote adaptive average pooling operators for matching sequence lengths. 
    We define
    \begin{equation}
    \bar{\mathbf{V}}^{T} = \mathcal{A}_{N_s}(\mathbf{V}^{T}), 
    \quad
    \bar{\mathbf{Q}}^{T} = \mathcal{A}_{M_s}(\mathbf{Q}^{T}),
    \quad
    \bar{\mathbf{p}}^{T} = \mathcal{A}_{H_p^S}(\hat{\mathbf{p}}^{T}),
    \quad
    \bar{\mathbf{w}}^{T} = \mathcal{A}_{H_w^S}(\hat{\mathbf{w}}^{T}).
    \end{equation}
    
    The waypoint, vision, query and waypoint-distillation losses are defined as:
    {\small
    \setlength{\jot}{1pt}
    \begin{align}
    L_{gt} &= 
    \frac{1}{H_p^S d_p}\left\|\hat{\mathbf{p}}^{S}-\mathbf{p}^{gt}\right\|_2^2
    +
    \frac{1}{H_w^S d_w}\left\|\hat{\mathbf{w}}^{S}-\mathbf{w}^{gt}\right\|_2^2, \\
    L_{vision} &=
    \frac{1}{N_sD_t}
    \left\|\phi_v(\mathbf{V}^{S})-\bar{\mathbf{V}}^{T}\right\|_2^2, \\
    L_{query} &=
    \frac{1}{M_sD_t}
    \left\|\phi_q(\mathbf{Q}^{S})-\bar{\mathbf{Q}}^{T}\right\|_2^2, \\
    L_{waypoint} &=
    \frac{1}{H_p^S d_p}\left\|\hat{\mathbf{p}}^{S}-\bar{\mathbf{p}}^{T}\right\|_1
    +
    \frac{1}{H_w^S d_w}\left\|\hat{\mathbf{w}}^{S}-\bar{\mathbf{w}}^{T}\right\|_1 .
    \end{align}
    }
    
    We further apply language supervision and language-logit distillation to preserve the teacher's language reasoning and explanation capability. 
    Under teacher guidance, the student generates logits at the target commentary positions. 
    Let $\boldsymbol{\ell}^{gt}$ denote the ground-truth commentary tokens and let $\mathcal{M}$ denote the set of valid target-text token positions. 
    The language ground-truth loss is defined as the token-normalized cross-entropy loss:
    \begin{equation}
    L_{\ell gt}
    =
    -\frac{1}{|\mathcal{M}|}
    \sum_{i \in \mathcal{M}}
    \log
    P^{S}_{i}(\ell^{gt}_{i}),
    \end{equation}
    where $P^{S}_{i}=\operatorname{softmax}(\mathbf{Z}_{\ell,i}^{S})$ is the student language distribution at token position $i$.
    
    In addition to ground-truth language supervision, we use teacher-student language-logit distillation. 
    Since the teacher and student language modules use the same tokenizer, their output logits are defined over a shared vocabulary, allowing direct KL-divergence-based distribution matching after aligning valid target-text positions. 
    Although the teacher and student models share the same tokenizer, their chat templates may introduce different prompt prefixes. 
    Therefore, we align the valid target-text spans on a per-sample basis before computing the KL divergence. 
    Specifically, we locate the first valid answer token in the student target mask, remove the teacher-side chat-template prefix and compute the maximum aligned token length from the remaining valid positions.
    
    For each aligned position $i \in \mathcal{M}$, we define
    $    \mathbf{P}_{i}^{T}
    =
    \operatorname{softmax}\left(\frac{\mathbf{Z}_{\ell,i}^{T}}{\tau}\right),
    \quad
    \mathbf{P}_{i}^{S}
    =
    \operatorname{softmax}\left(\frac{\mathbf{Z}_{\ell,i}^{S}}{\tau}\right),$

    where $\tau$ is the temperature hyperparameter. 
    The language distillation loss is defined as the token-normalized KL divergence:
    \begin{equation}
    L_{kl}
    =
    \frac{\tau^2}{|\mathcal{M}|}
    \sum_{i \in \mathcal{M}}
    D_{\mathrm{KL}}
    \left(
    \mathbf{P}_{i}^{T}
    \;\|\;
    \mathbf{P}_{i}^{S}
    \right).
    \end{equation}
    
    This global token-level normalization weighs each valid token equally across the batch, and hence longer commentary sequences naturally contribute proportionally more valid-token supervision.
    
    The final training objective is decoupled into a driving loss and a language loss, facilitating a two-stage training scheme. In the first stage, the driving loss $L_{\mathrm{driving}}$ is employed to optimize the student's driving capabilities; the driving branch is subsequently frozen. In the second stage, the model is trained exclusively using the language loss $L_{\mathrm{language}}$ to specialize in explanation generation. The objective terms are formulated as follows:
    \begin{equation}
    L_{\mathrm{driving}} = \lambda_g L_{gt} + \lambda_v L_{vision} + \lambda_q L_{query} + \lambda_w L_{waypoint}
    \end{equation}
    \begin{equation}
    L_{\mathrm{language}} = \lambda_{\ell gt} L_{\ell gt} + \lambda_{kl} L_{kl}
    \end{equation}
    where $\lambda_g$, $\lambda_v$, $\lambda_q$, $\lambda_w$, $\lambda_{\ell gt}$ and $\lambda_{kl}$ control the relative importance of waypoint ground-truth supervision, vision feature distillation, query feature distillation, waypoint prediction distillation, language ground-truth supervision and language-logit distillation, respectively.
\vspace{-1mm}
    \subsection{On-Policy Language Fine-Tuning}
    \label{on-policy}
\vspace{-1mm}
    The language-logit distillation loss in Sec. \ref{loss_des} is computed under teacher supervision, while inferencing uses autoregressive generation from the student's previous tokens. To reduce this shift, we apply on-policy language fine-tuning after offline distillation. The driving branch is frozen to provide stable visual and intermediate features and the student language branch generates rollout tokens using a standard greedy decoding strategy. These tokens are appended to the prompt and evaluated by the frozen teacher to obtain teacher logits at the rollout positions; the student re-evaluates the same sequence and KL divergence is minimized over these on-policy tokens. This directly supervises the student's generated text, improving post-hoc explanation consistency while preserving the low-latency driving branch.

\vspace{-2mm}
\section{Experimental Evaluation of RT-VLA}
\label{sec:result}
\vspace{-1mm}

\subsection{Experimental Settings}
\vspace{-1mm}
We evaluate the distilled student model on the same Bench2Drive split used by the SimLingo family, consisting of 220 routes in CARLA v0.9.15. 
The model is trained on SimLingo's training data, which covers 15 driving scenario categories, with 5\% of the data reserved for validation. 
For evaluation, we report three metrics: the Bench2Drive driving score (DS), which combines route completion with infraction penalties; average inference time per frame, which measures real-time deployment efficiency; and language commentary quality. All training and inference tests are conducted on an NVIDIA A100 40GB GPU, with training taking 48 GPU hours. For RT-VLA, we separately report the vision-only driving mode and the vision-language mode to distinguish real-time control latency from the additional cost of language reasoning. 
Commentary quality is evaluated by comparing generated and ground-truth commentary using DeepSeek-V4-Flash, which assesses semantic correctness, driving relevance and explanatory quality. 
All language evaluations use the same judge and evaluation protocol to ensure consistent comparison across models with commentary outputs.
    \vspace{-1mm}
    \subsection{Driving Performance}
    \vspace{-1mm}
    Table~\ref{tab:driving_performance} compares RT-VLA performance with that of the SimLingo family on the Bench2Drive split.
    
    \begin{table}[t]
    \centering
    \caption{Driving performance, inference efficiency and commentary quality on Bench2Drive. DS denotes driving score. Commentary quality is evaluated using DeepSeek-V4-Flash.}
    \label{tab:driving_performance}
    \small
    \setlength{\tabcolsep}{3pt}
    \begin{tabular}{lcccc}
    \toprule
    Method & Modality & DS $\uparrow$ & Time / Frame $\downarrow$ & Commentary Quality $\uparrow$ \\
    \midrule
    SimLingo & Vision+Language & 85.07 & 1544.34 ms & \textbf{51.8} \\
    SimLingo-BASE & Vision & \textbf{85.94} & 41.11 ms & -- \\
    \midrule
    \multirow{2}{*}{RT-VLA (Ours)} & Vision+Language & 85.19 & \textbf{196 ms} & 50.9 \\
    & Vision & 85.19 & \textbf{34.48 ms} & -- \\
    \bottomrule
    \end{tabular}
    \end{table}
    
    \textbf{Comparable closed-loop driving performance.}
    RT-VLA achieves a driving score of 85.19, which is comparable to both SimLingo and SimLingo-BASE. 
    Specifically, RT-VLA slightly outperforms SimLingo, which obtains a driving score of 85.07, while remaining close to SimLingo-BASE, which obtains 85.94. 
    Although SimLingo-BASE achieves the highest driving score among the compared models, the gap between RT-VLA and SimLingo-BASE is only 0.75 points. Notably, SimLingo-BASE does not provide any language reasoning capability. This outcome indicates that RT-VLA preserves the closed-loop driving capability of the larger teacher model while additionally retaining language-based reasoning ability in a lightweight student architecture.
    
    \textbf{Improved inference efficiency.}
    RT-VLA achieves the lowest inference time among the compared models, requiring 34.48 ms per frame on a single NVIDIA A100 40GB GPU. 
    Compared with SimLingo, RT-VLA reduces inference time from 1544.34 ms to 34.48 ms, corresponding to approximately a $44.8\times$ speedup. 
    With the language branch enabled, RT-VLA still reduces inference time from 1544.34 ms to 196 ms compared with SimLingo, corresponding to approximately a $7.9\times$ speedup.
    Compared with SimLingo-BASE, RT-VLA reduces inference time from 41.11 ms to 34.48 ms, corresponding to approximately a $19\%$ speedup. 
    These results show that RT-VLA provides a favorable performance-efficiency trade-off, maintaining competitive driving performance while substantially improving inference efficiency.

    \textbf{Preserved language commentary capability.}
    As an intended outcome, in addition to closed-loop driving performance, RT-VLA retains language commentary capability after language distillation and on-policy fine-tuning. 
    Using DeepSeek-V4-Flash as the evaluator, RT-VLA obtains a commentary score of 50.9, compared with 51.8 for the full SimLingo model. 
    Although RT-VLA remains slightly below the full teacher model in commentary quality, the gap is only 0.9 points despite the substantial reduction in inference latency. 
    This result demonstrates that RT-VLA can preserve most of the teacher's language-based explanation capability while remaining suitable for real-time driving under real-world conditions.
    
\vspace{-1mm}
\subsection{Effect of Distillation and On-Policy Fine-Tuning}
\vspace{-1mm}
    \begin{table}[t]
        \centering
        \caption{Effect of distillation and on-policy language fine-tuning on driving performance and language commentary quality. Commentary quality is evaluated using DeepSeek-V4-Flash.}
        \label{tab:distillation_ablation}
        \small
        \setlength{\tabcolsep}{4pt}
        \begin{tabular}{lcc}
        \toprule
        Method & DS $\uparrow$ & Commentary $\uparrow$ \\
        \midrule
        RT-VLA w/o distillation & 34.05 & 44.6 \\
        RT-VLA w/ distillation & \textbf{85.17} & 47.0 \\
        RT-VLA w/ on-policy language FT & \textbf{85.17} & \textbf{50.9} \\
        \bottomrule
        \end{tabular}
    \end{table}
    
    We analyze the impact of distillation by comparing student models trained with and without teacher supervision. As shown in Table~\ref{tab:distillation_ablation}, the student trained without distillation achieves a driving score of only 34.05, indicating direct supervised training is insufficient for recovering strong closed-loop driving behavior with a lightweight architecture. After applying multi-level distillation, the driving score increases to 85.17, corresponding to an absolute improvement of 51.12 points and approximately a $2.5\times$ relative improvement. The teacher provides structured guidance through final waypoint predictions and intermediate visual and query representations, helping the student recover stronger driving policies. Distillation and on-policy language fine-tuning also improve commentary quality, increasing the DeepSeek-based score from 44.6 to 50.9. This indicates that teacher supervision helps the student preserve both driving behavior and language-related reasoning capability.
    \vspace{-5mm}
    \subsection{Qualitative Observations}
\vspace{-1mm}
    \begin{figure*}[t]
      \centering
    
      \begin{subfigure}[t]{0.495\textwidth}
        \centering
        \includegraphics[width=\linewidth, trim=5 5 5 5, clip]{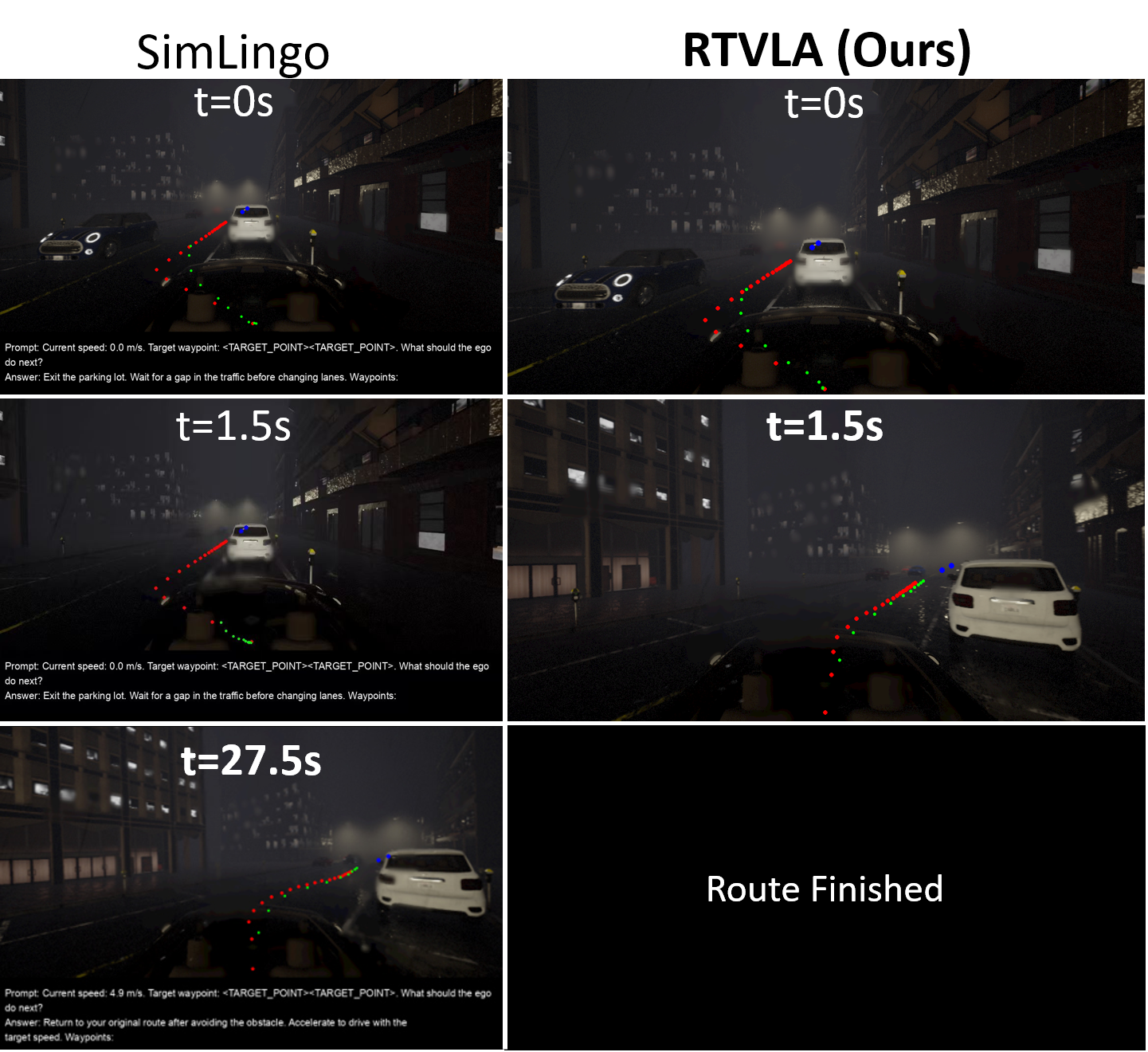}
        \caption{Parking-exit scenario. RT-VLA reaches a comparable obstacle-avoidance stage earlier than SimLingo.}
        \label{fig:parking_exit}
      \end{subfigure}
      \hspace{-0.01\textwidth}
      \begin{subfigure}[t]{0.495\textwidth}
        \centering
        \includegraphics[width=\linewidth, trim=5 5 5 5, clip]{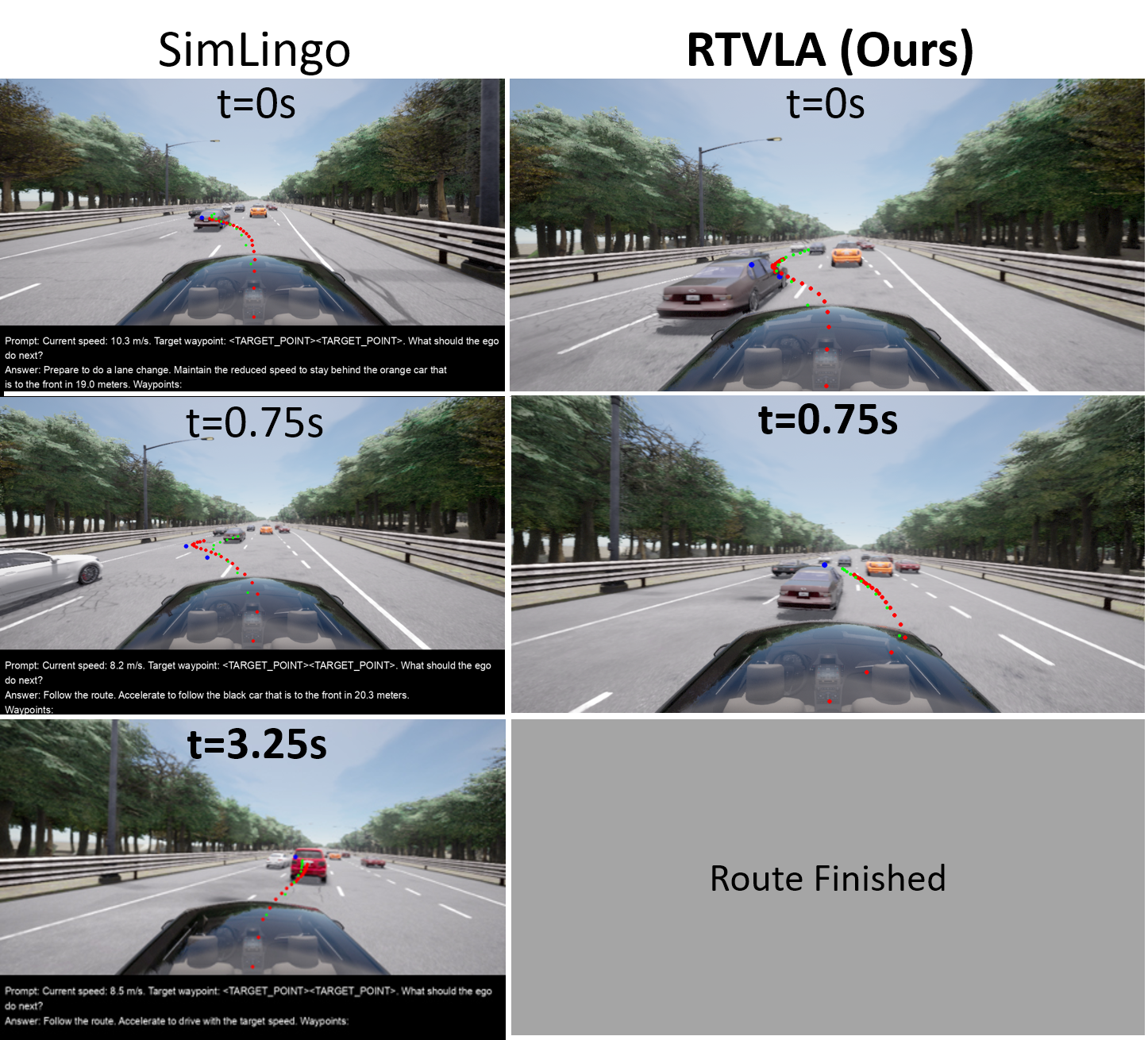}
        \caption{Highway lane-change scenario. RT-VLA reacts earlier and completes the lane-change maneuver faster.}
        \label{fig:change_lane}
      \end{subfigure}
    \vspace{-1mm}
      \caption{
      Qualitative comparison between SimLingo and RT-VLA in closed-loop driving scenarios.
      }
      \label{fig:qualitative_results}
    \end{figure*}

    We further visualize the closed-loop behavior of RT-VLA and compare it with SimLingo in two scenarios, as shown in Fig.~\ref{fig:qualitative_results}. 
    We measure responsiveness from the initial observation, denoted as $t=0$s. 
    In the parking-exit scenario, both models initially observe a vehicle ahead and generate waypoints that avoid direct collision. 
    However, RT-VLA reaches a comparable obstacle-avoidance stage by $t=1.5$s, while SimLingo remains near the initial waiting position and only reaches a similar maneuver stage later at $t=27.5$s. At this stage, many other vehicles have passed the ego vehicle and due to the high inference latency of SimLingo, it has not found a viable solution to merge into traffic safely. However, RT-VLA has merged much earlier and finished the route by this time. This suggests that the reduced inference latency of RT-VLA significantly improves ego-vehicle responsiveness, especially in dense and constrained urban scenes.
    
    In the highway lane-change scenario, both models observe nearby vehicles and generate trajectories for lane adjustment at $t=0$s. 
    RT-VLA progresses through the lane-change maneuver earlier and completes the route by $t=3.25$s, whereas SimLingo is still executing the maneuver at that elapsed time. 
    These qualitative examples support the quantitative results that RT-VLA preserves the main closed-loop driving behavior of SimLingo while enabling faster action updates.
\vspace{-1mm}
\subsection{Offline Language Explanation}
\begin{figure*}[t]
    \centering
    \begin{subfigure}[t]{0.32\textwidth}
        \centering
        \includegraphics[width=\textwidth]{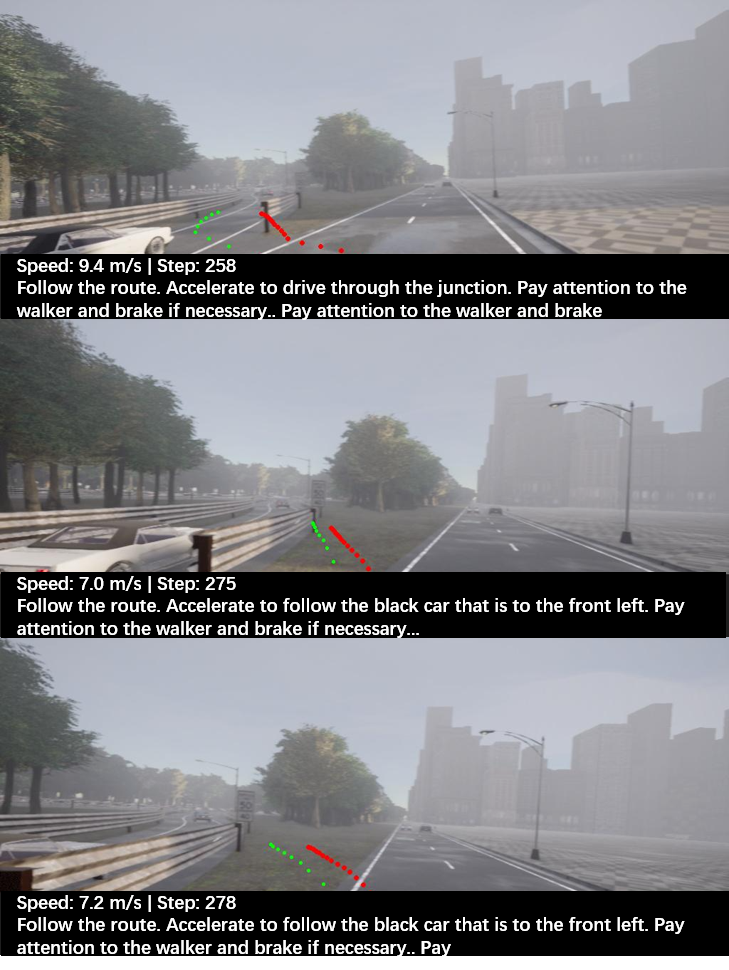}
        \caption{Outside-route scenario. The retrieved comments suggest that the model attempts to follow the black car ahead, but fails to recognize the route split, leading to a deviation from the designated route.}
        \label{fig:outside_route}
    \end{subfigure}
    \hfill
    \begin{subfigure}[t]{0.32\textwidth}
        \centering
        \includegraphics[width=\textwidth]{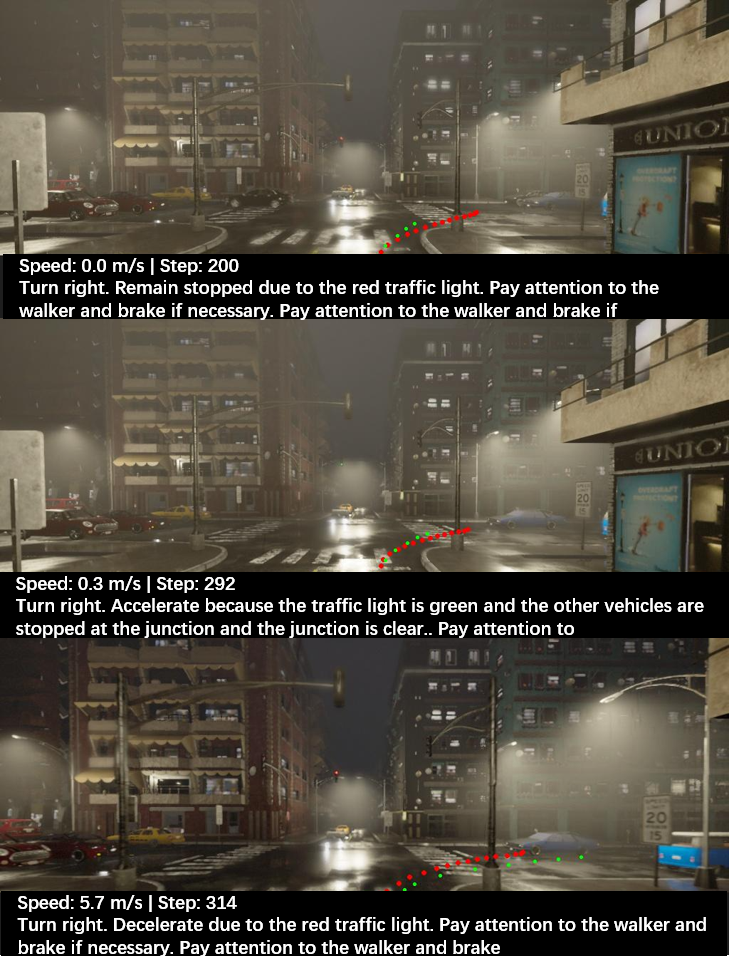}
        \caption{Traffic-light scenario. The retrieved comments indicate that the model initially stops for the red light, but later proceeds with the turn after the light turns red again, resulting in a red-light violation.}
        \label{fig:traffic_light}
    \end{subfigure}
    \hfill
    \begin{subfigure}[t]{0.32\textwidth}
        \centering
        \includegraphics[width=\textwidth]{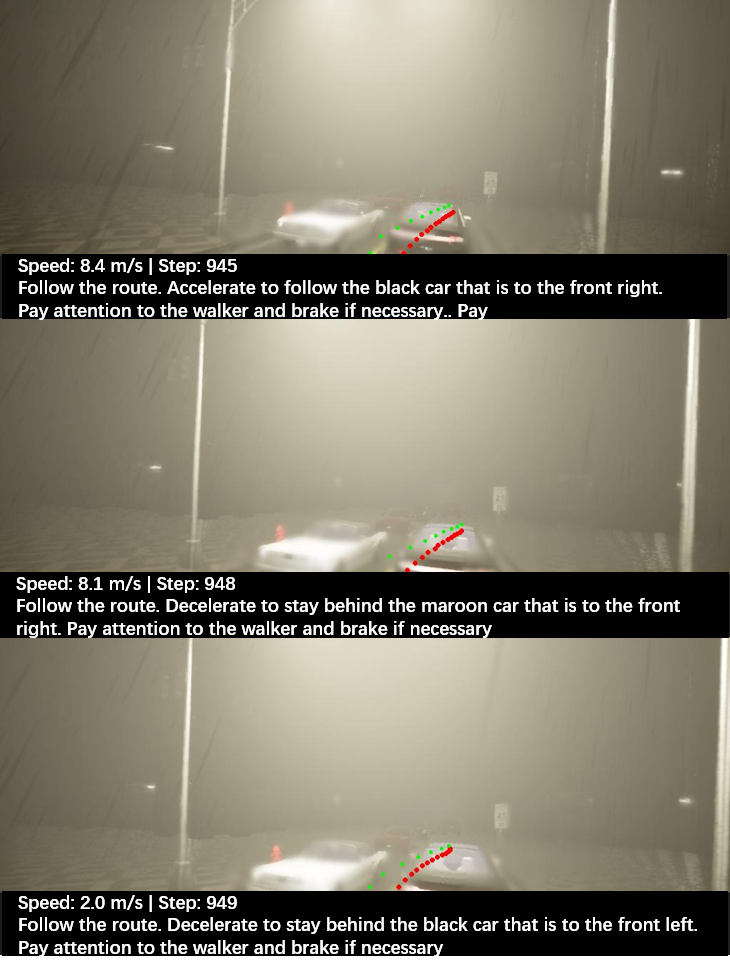}
        \caption{Collision scenario. The retrieved comments show that the model tries to follow the vehicle ahead, then decelerates as it approaches, but fails to stop in time and results in a rear-end collision.}
        \label{fig:collision}
    \end{subfigure}
    \caption{Offline language explanation from logged driving observations.}
    \label{fig:offline_explanation}
\end{figure*}

\vspace{-1mm}
We next demonstrate the offline language explainability of RT-VLA by retrieving language comments from logged driving observations in safety-critical scenarios. 
During real-time closed-loop control, the lightweight driving branch predicts waypoints without invoking the full language reasoning module, avoiding additional inference latency. 
If an explanation is needed after a failure, the set of logged frames preceding the failure can be passed to the language reasoning branch offline to generate post-hoc comments describing the driving context, the intended maneuver and potential risk factors. 
Figure~\ref{fig:offline_explanation} shows three representative cases: deviating from the designated route, running a red light and colliding with vehicles ahead. 
In each case, the retrieved comments describe the scene and model behavior, supporting offline reasoning without interfering with real-time control. This language explainability can help identify the failure modes of VLAs for autonomous driving, informing the development of safer E2E models.

\vspace{-2mm}
\section{Limitations}
\vspace{-1mm}
Although RT-VLA achieves a favorable trade-off between closed-loop driving performance, language capability and inference efficiency, several limitations remain. First, RT-VLA cannot fully eliminate safety-critical failures such as collisions, since it is trained through supervision and distillation rather than explicit safety-constrained optimization. Second, RT-VLA is a camera-only framework without LiDAR or other geometric sensors, which may limit its robustness under adverse environmental conditions such as heavy rain, fog, low illumination or glare. Third, the model inherits the limitations of the teacher model and the simulation-based training environment. Therefore, systematic teacher errors, long-tail scenarios and real-world domain shifts may affect its reliability. 

\vspace{-2mm}
\section{Conclusion}
\vspace{-1mm}
In this work, we presented RT-VLA, a lightweight distilled model for end-to-end autonomous driving. Motivated by the critical need to reduce high inference latency of VLA-based driving models, RT-VLA transfers the driving and reasoning capabilities of a larger SimLingo teacher model into a compact student model through multi-level supervised distillation. By aligning visual features, query representations, waypoint predictions and language logits, and applying on-policy language fine-tuning, RT-VLA preserves competitive closed-loop driving performance while substantially reducing inference cost. It also supports offline post-hoc explanation for safety-critical incidents without adding latency to real-time on-road control. Our experiments demonstrate strong efficiency gains while retaining driving capability and explainability. Our future work will explore stronger safety-aware objectives, multimodal sensor fusion and real-world validation to further improve robustness and deployment readiness.




\bibliography{example}  

\end{document}